\documentclass[letterpaper, 10pt, conference]{style/ieeeconf}    

\makeatletter
\let\NAT@parse\undefined
\makeatother

\usepackage{dblfloatfix}

\usepackage[numbers,sectionbib,sort&compress]{natbib}
\usepackage{bm}
\usepackage{gensymb}
\usepackage{graphicx}
\usepackage{amsmath}
\usepackage{amssymb}
\usepackage{algorithm}
\usepackage[noend]{algpseudocode}
\usepackage{subcaption}
\usepackage{amsfonts}
\usepackage{siunitx}
\usepackage{booktabs}
\usepackage{makecell}
\usepackage{multirow}
\usepackage{upgreek}
\usepackage[font=small]{caption}
\usepackage[export]{adjustbox}
\usepackage{tikz}
\usepackage{sidecap} \sidecaptionvpos{figure}{c}
\DeclareMathOperator{\Tr}{Tr}
\DeclareMathOperator*{\argmax}{argmax}

\usepackage{hyperref}
\hypersetup{colorlinks,breaklinks,
linkcolor=[rgb]{0.5,0.,0.},
citecolor=[rgb]{0.000,0.427,0.173},
urlcolor=[rgb]{0.031,0.318,0.612}}

\usepackage[nameinlink, capitalize]{cleveref}
\usepackage{color}
\definecolor{CommentPink}{rgb}{1,0.2,0.5}
\definecolor{CommentBlue}{rgb}{0,0,1}
\definecolor{CommentGreen}{rgb}{0,1,0}

\Crefname{section}{Sec.}{Sec.}

\usepackage[printonlyused,withpage,nolist,nohyperlinks]{acronym}
\begin{acronym}

\acro{2D}{two-dimensional}

\acro{3D}{three-dimensional}

\acro{AHRS}{attitude and heading reference system}

\acro{AUV}{autonomous underwater vehicle}

\acro{CPP}{Chinese Postman Problem}

\acro{DoF}{degree of freedom}
\acrodefplural{DoF}[DoFs]{degrees of freedom}

\acro{DVL}{Doppler velocity log}

\acro{FSM}{finite state machine}

\acro{IMU}{inertial measurement unit}

\acro{LBL}{Long Baseline}

\acro{MCM}{mine countermeasures}

\acro{MDP}{Markov decision process}
\acrodefplural{MDP}[MDPs]{Markov decision processes}

\acro{POMDP}{Partially Observable Markov Decision Process}
\acrodefplural{POMDP}[POMDPs]{Partially Observable Markov Decision Processes}

\acro{PRM}{Probabilistic Roadmap}
\acrodefplural{PRM}[PRM]{Probabilistic Roadmaps}

\acro{ROI}{region of interest}
\acrodefplural{ROI}[ROIs]{regions of interest}

\acro{ROS}{Robot Operating System}

\acro{ROV}{remotely operated vehicle}

\acro{RRT}{Rapidly-exploring Random Tree}
\acrodefplural{RRT}[RRTs]{Rapidly-exploring Random Trees}

\acro{SLAM}{Simultaneous Localisation and Mapping}

\acro{SSE}{sum of squared errors}

\acro{STOMP}{Stochastic Trajectory Optimization for Motion Planning}

\acro{TRN}{Terrain-Relative Navigation}

\acro{UAV}{unmanned aerial vehicle}

\acro{USBL}{Ultra-Short Baseline}

\acro{IPP}{informative path planning}

\acro{FoV}{field of view}
\acrodefplural{FoV}[FoVs]{fields of view}

\acro{CDF}{cumulative distribution function}

\acro{ML}{maximum likelihood}

\acro{RMSE}{Root Mean Squared Error}
\acro{MLL}{Mean Log Loss}

\acro{GP}{Gaussian Process}
\acrodefplural{GP}[GPs]{Gaussian Processes}

\acro{KF}{Kalman Filter}

\acro{IP}{Interior Point}
\acro{BO}{Bayesian Optimization}

\acro{SE}{squared exponential}

\acro{UI}{uncertain input}

\acro{MCL}{Monte Carlo Localisation}
\acro{AMCL}{Adaptive Monte Carlo Localisation}

\acro{SSIM}{Structural Similarity Index}

\acro{MAE}{Mean Absolute Error}
\acro{RMSE}{Root Mean Squared Error}
\acro{AUSE}{Area Under the Sparsification Error curve}

\end{acronym}

\IEEEoverridecommandlockouts                           
\title{\LARGE \bf
Adaptive Informative Path Planning Using \\ Deep Reinforcement Learning for UAV-based Active Sensing}
\author{Julius Rückin, Liren Jin, Marija Popovi\'{c}
\thanks{This work was funded by the Deutsche Forschungsgemeinschaft (DFG, German Research Foundation) under Germany's Excellence Strategy - EXC 2070 – 390732324.
Authors are with the Cluster of Excellence PhenoRob, Institute of Geodesy and Geoinformation, University of Bonn. Corresponding: \texttt{jrueckin@uni-bonn.de}.}%
}

\begin{document}

\maketitle

\begin{abstract}
Aerial robots are increasingly being utilized for environmental monitoring and exploration.
However, a key challenge is efficiently planning paths to maximize the information value of acquired data as an initially unknown environment is explored.
To address this, we propose a new approach for informative path planning based on deep reinforcement learning (RL).
Combining recent advances in RL and robotic applications, our method combines tree search with an offline-learned neural network predicting informative sensing actions.
We introduce several components making our approach applicable for robotic tasks with high-dimensional state and large action spaces.
By deploying the trained network during a mission, our method enables sample-efficient online replanning on platforms with limited computational resources.
Simulations show that our approach performs on par with existing methods while reducing runtime by $8-10\times$.
We validate its performance using real-world surface temperature data.
\end{abstract}

\section{Introduction} \label{S:introduction}

Recent years have seen an increasing usage of autonomous robots in a variety of data collection applications, including environmental monitoring \citep{popovic2020informative,hitz2017adaptive,hollinger2014sampling,dunbabin2012robots,lelong2008assessment}, exploration \citep{doherty2007uav}, and inspection \citep{Galceran2013}. In many tasks, these systems promise a more flexible, safe, and economic solution compared to traditional manual or static sampling methods \citep{dunbabin2012robots, vivaldini2019uav}. However, to fully exploit their potential, a key challenge is developing algorithms for \textit{active sensing}, where the objective is to plan paths for efficient data gathering subject to finite computational and sensing resources, such as energy, time, or travel distance.

This paper examines the task of active sensing using an unmanned aerial vehicle (UAV) in terrain monitoring scenarios. Our goal is to map an a priori unknown nonhomogeneous 2D scalar field, e.g. of temperature, humidity, etc., on the terrain using measurements taken by an on-board sensor. In similar setups, most practical systems rely on precomputed paths for data collection, e.g. coverage planning \cite{Galceran2013}. However, such approaches assume a uniform distribution of measurement information value in the environment and hence do not allow for \textit{adaptivity}, i.e. closely inspecting regions of interest, such as hotspots \citep{hitz2017adaptive, popovic2020informative} or anomalies \citep{blanchard2020informative}, as they are discovered. Our motivation is to find information-rich paths targeting these areas by performing efficient online adaptive replanning on computationally constrained platforms.

Several \textit{informative path planning (IPP)} approaches for active sensing have been proposed \citep{popovic2020informative,choudhury2020adaptive,hitz2017adaptive,hollinger2014sampling,vivaldini2019uav}, which enable adjusting decision-making based on observed data. However, scaling these methods to large problem spaces remains an open challenge. The main computational bottleneck in IPP is the predictive replanning step, since multiple future measurements must be simulated when evaluating next candidate actions. Previous studies have tackled this by discretizing the action space, e.g. sparse graphs \citep{choudhury2020adaptive,popovic2020localisation}; however, such simplifications sacrifice on the quality of predictive plans. An alternative paradigm is to use reinforcement learning (RL) to learn data gathering actions. Though emerging works in RL for IPP demonstrate promising results \cite{viseras2019deepig, chen2020autonomous}, they have been limited to small 2D action spaces, and adaptive planning to map environments with spatial correlations and large 3D action spaces has not yet been investigated.

 
 \begin{figure}[!t]
    \centering
    \includegraphics[width=0.215\textwidth]{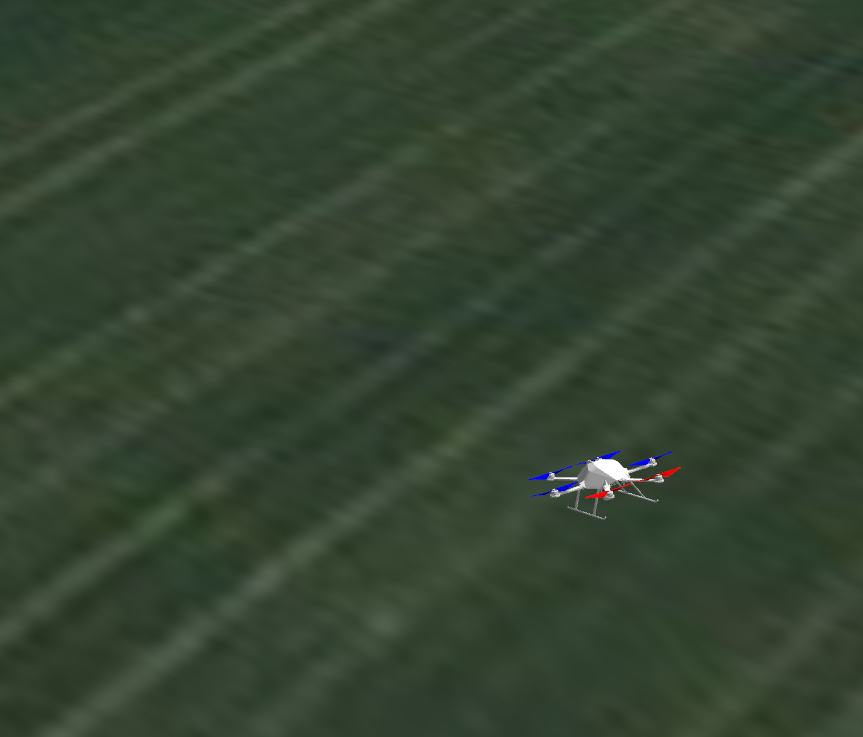}
    \includegraphics[width=0.207\textwidth]{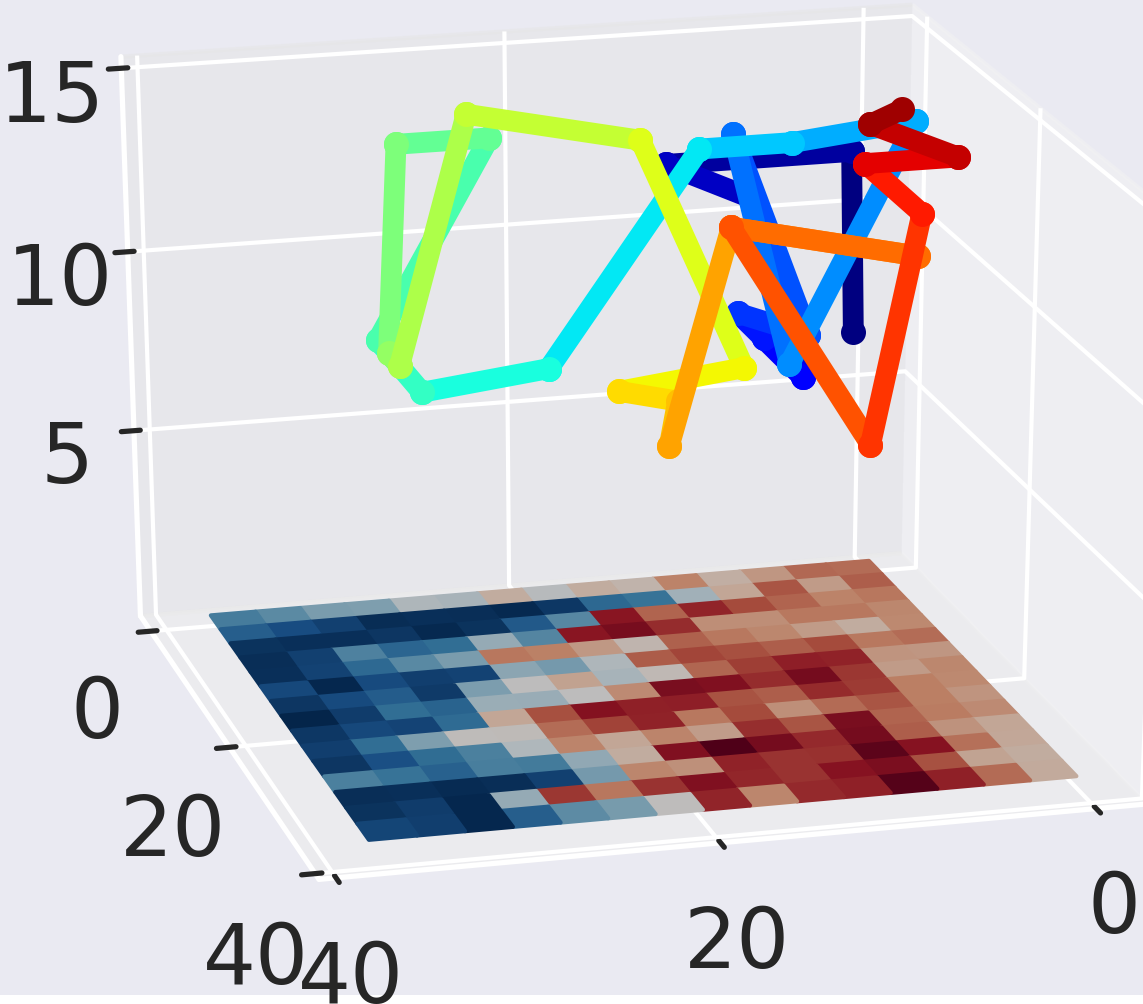}
    \caption{\textbf{Our RL-based IPP approach applied in a temperature mapping scenario.} (Left) Our simulation setup using real-world field temperature data. (Right) The temperature variable is projected to the surface. Our approach \textit{adaptively} plans a UAV's path (evolving over time from blue to red) focusing on hotter regions (red).} \label{F:teaser}
\end{figure}

To address this, we propose a new RL-based IPP framework suitable for UAV-based active sensing. Inspired by recent advances in RL \citep{silver2017mastering, silver2018general}, our method combines Monte Carlo tree search (MCTS) with a convolutional neural network (CNN) to learn information-rich actions in adaptive data gathering missions. Since active sensing tasks are typically expensive to simulate, our approach caters for training in low data regimes. By replacing the computational burden of predictive planning with simple tree search, we achieve efficient online replanning, which is critical for deployment on mobile robots (\cref{F:teaser}).
The contributions of this work are:
\begin{enumerate}
    \item A new deep RL algorithm for robotic planning applications that supports continuous high-dimensional state spaces, large action spaces, and data-efficient training.
    \item The integration of our RL algorithm in an IPP framework for UAV-based terrain monitoring.
    \item The validation of our approach in an ablation study and evaluations against benchmarks using synthetic and real-world data showcasing its performance.
\end{enumerate}
We open-source our framework for usage by the community.\footnote{\texttt{github.com/dmar-bonn/ipp-rl}}

\section{Related Work} \label{S:related_work}

Our work lies at the intersection of IPP for active sensing, MCTS planning methods, and recent advances in RL.

IPP methods are gaining rapid traction in many active sensing applications \citep{hitz2017adaptive, popovic2020informative, hollinger2014sampling,choudhury2020adaptive}.
In this area of study, our work focuses on strategies with \textit{adaptive} online replanning capabilities, which allow the targeted monitoring of regions of interest, e.g. hotspots or abnormal areas, in \textit{a priori unknown} environments \cite{blanchard2020informative}. Some methods focus on discrete action spaces defined by sparse graphs of permissible actions \citep{choudhury2020adaptive,popovic2020localisation}. However, these simplifications are not applicable as the distribution of target regions requires online decision-making as they are discovered. Our proposed algorithm reasons about a discrete action space magnitudes larger while ensuring online computability. In terms of planning strategy, IPP algorithms can be classified into combinatorial \cite{ko1995exact, binney2012branch}, sampling-based \cite{hollinger2014sampling, choudhury2020adaptive}, and optimization-based approaches \cite{popovic2020informative, hitz2017adaptive}. Combinatorial methods exhaustively query the search space. Thus, they cannot plan online in large search spaces, which makes them impractical for adaptive replanning.

Continuous-space sampling-based planners generate informative robot trajectories by sampling candidate actions while guaranteeing probabilistically asymptotic optimality \citep{hollinger2014sampling, vivaldini2019uav}. 
However, their sample-efficiency is typically low for planning with more complex objectives and larger action spaces since many measurements need to be forward-simulated to find promising paths in the problem space \cite{popovic2020informative, omidvar2010comparative}.
In our particular setup, considering spatial correlations in a terrain over many candidate regions leads to a complex and expensive-to-evaluate information criterion.
In similar scenarios, several works have investigated global optimization, e.g. evolutionary algorithms \cite{hitz2017adaptive, popovic2020informative} or Bayesian Optimization \cite{vivaldini2019uav}, to enhance planning efficiency.
Although these approaches deliver high-quality paths \cite{popovic2020informative, popovic2020localisation}, using them for online decision-making is still computationally expensive when reasoning about many spatially correlated candidate future measurements.

In robotics, RL algorithms are increasingly being utilized for search and rescue \cite{niroui2019deep}, information gathering \cite{viseras2019deepig}, and exploration of unknown environments \cite{chen2020autonomous}. Although emerging works show promising performance, RL-based approaches have not yet been investigated for online adaptive IPP, instead being mostly restricted to environment exploration tasks. To address this gap, we propose the first RL approach for \textit{adaptively} planning informative paths online over spatially correlated terrains with large action spaces.

Another well-studied planning paradigm is MCTS \cite{browne2012survey, chaslot2008monte}. Recently, MCTS extensions were proposed for large action spaces \cite{sunberg2018online} and partially observable environments \cite{silver2010monte}. \citet{choudhury2020adaptive} applied a variant of MCTS to obtain long-horizon, anytime solutions in adaptive IPP problems. However, these online methods are restricted to small action spaces and spatially uncorrelated environments.

Inspired by recent advances in RL \cite{silver2018general, silver2017mastering}, our RL-based algorithm bypasses computationally expensive MCTS rollouts and sample-inefficient action selections with a learned value function and an action policy, respectively. We extend the AlphaZero algorithm by applying it to robotics tasks with limited computational budget and low data regimes. Related to our approach are RL methods which address the exploration-exploitation trade-off at train time \citep{schulman2017proximal, haarnoja2018soft}. However, our algorithm explicitly balances exploration and exploitation at deploy time by combining a learned policy with MCTS sampling-based planning in large action spaces.

\section{Background} \label{S:background}
We begin by briefly describing the general active sensing problem and specifying the terrain monitoring scenario used to develop our RL-based IPP approach.

\subsection{Problem Formulation} \label{SS:problem_formulation}

The general active sensing problem aims to maximize an information-theoretic criterion $\mathrm{I}(\cdot)$ over a set of action sequences $\Psi$, i.e. robot trajectories: 
\begin{equation} \label{eq:1}
    \psi^* = \argmax_{\psi \in \Psi} \frac{\mathrm{I}(\psi)}{\mathrm{C}(\psi)},\, \text{s.t. } \mathrm{C}(\psi) \leq \mathrm{B},
\end{equation}
where $\mathrm{C}: \Psi \to \mathbb{R}^{+}$ maps an action sequence to its associated execution cost, $\mathrm{B} \in \mathbb{R}^{+}$ is the robot's budget limit, e.g. time or energy, and $\mathrm{I}: \Psi \to \mathbb{R}^{+}$ is the information criterion, computed from the new sensor measurements obtained by executing $\psi$.

This work focuses on the scenario of monitoring a terrain using a UAV equipped with a camera. Specifically, in this case, the costs $\mathrm{C}(\psi)$ of an action sequence $\psi = (\psi_1, \ldots, \psi_n)$ of length $n$ are defined by the total flight time:
\begin{equation} \label{eq:2}
    \mathrm{C}(\psi) = \sum_{i=1}^{n-1} \mathrm{c(\psi_i, \psi_{i+1})},
\end{equation}
where $\psi_{i} \in \mathbb{R}^3$ is a 3D measurement position above the terrain the image is registered from. $c: \mathbb{R}^3 \times \mathbb{R}^3 \to \mathbb{R}^{+}$ computes the flight time costs between measurement positions by a constant acceleration-deceleration $\pm u_{a}$ model with maximum speed $u_{v}$.

\subsection{Terrain Mapping} \label{SS:mapping}

We leverage the method of \citet{popovic2020informative} for efficient probabilistic mapping of the terrain. The terrain $\xi \subset \mathbb{R}^{2}$ is discretized by a grid map $\mathcal{X}$ while the mapped target variable, e.g. temperature, is a scalar field $\zeta: \xi \to \mathbb{R}$. The prior map distribution $p(\zeta\,|\,\xi) \sim \mathcal{N}(\bm{\mu^{-}}, \bm{P^{-}})$ is given by a Gaussian Process (GP) defined by a prior mean vector $\bm{\mu^{-}}$ and covariance matrix $\bm{P^{-}}$. At mission time, a Kalman Filter $\mathcal{KF}(\bm{\mu^{-}}, \bm{P^{-}}, \bm{z}, \psi_i) = \bm{\mu^{+}}, \bm{P^{+}}$ is used to sequentially fuse data $\bm{z} \in \mathbb{R}^m$ observed at a measurement location $\psi_i$ with the last iteration's map belief $p(\zeta\,|\,\xi) \sim \mathcal{N}(\bm{\mu^{-}}, \bm{P^{-}})$ in order to obtain the posterior map mean $\bm{\mu^{+}}$, and covariance $\bm{P^{+}}$. For further details, the reader is referred to \citep{popovic2020informative}.

\subsection{Utility Definition for Adaptive IPP} \label{SS:utility}

In \cref{eq:1}, we define the A-optimal information criterion associated with an action sequence of $n$ measurements \cite{sim2005global}:
\begin{equation} \label{eq:3}
    \mathrm{I}(\psi) = \sum_{i=1}^n \Tr(\bm{P^-}) - \Tr(\bm{P^{+}}),
\end{equation}
%
where $\bm{P^-}$ and $\bm{P^+}$ are obtained before and after applying $\mathcal{KF}$ to the measurements observed along $\psi_{i}$, respectively.

We study an active sensing task where the goal is to gather terrain areas with higher values of the target variable $\zeta$, e.g. high temperature. This scenario requires online replanning to focus on mapping these areas of interest $\mathcal{X_{\mathrm{I}}}$ as they are discovered, and is thus a relevant problem setup for our new efficient RL-based IPP strategy. We utilize confidence-based level sets to define $\mathcal{X_{\mathrm{I}}}$ \cite{gotovos2013}:
\begin{equation} \label{eq:4}
    \mathcal{X_{\mathrm{I}}} = \{x_i \in \mathcal{X} ~ \vert ~ \bm{\mu}^{-}_{i} + \beta \bm{P}^{-}_{i, i} \geq \mu_{th}\},
\end{equation}
where $\bm{\mu}^{-}_{i}$ and $\bm{P}^{-}_{i, i}$ are the mean and variance of grid cell $x_{i}$. $\beta, ~ \mu_{th} \in \mathbb{R}^{+}$ are a user-defined confidence interval width and threshold, respectively.
Consequently, we restrict $\bm{P^-}$ and $\bm{P^+}$ in \cref{eq:3} to the grid cells $x_{i} \in \mathcal{X_{\mathrm{I}}}$ as defined by \cref{eq:4}, noted as $\bm{P^-}_{\mathcal{X_{\mathrm{I}}}}$ and $\bm{P^+}_{\mathcal{X_{\mathrm{I}}}}$ respectively. 



\section{Approach} \label{S:approach}

This section presents our new RL-based IPP approach for active sensing. As shown in \cref{F:overview}, we iteratively train a CNN on diverse simulated terrain monitoring scenarios to learn the most informative data gathering actions. The trained CNN is leveraged during a mission to achieve fast online replanning. 

\subsection{Connection between IPP \& RL} \label{SS:ipp_rl_connection}

We first cast the general IPP problem from \cref{S:background} in a RL setting. The value $V(s)$ of a state $s$ is defined as $V(s) = r(s, a, s') + \gamma V(s')$, where $\gamma \in [0,1]$, and $s'$ is the successor state when choosing a next action $\psi_{i+1} = a \sim \bm{\pi}(s)$ according to the policy $\bm{\pi}(\cdot)$. A state $s$ is defined by $s = (s_{m}, a^{-})$, where $s_{m} \sim \mathcal{N}(\bm{\mu^{-}}, \bm{P^{-}})$ is the current map state, and $a^{-}$ is the previously executed action, i.e. the current UAV position. Consequently, $s' = (s'_{m}, a)$ is defined by $s'_{m} \sim \mathcal{N}(\bm{\mu^{+}}, \bm{P^{+}})$. In our work, the 3D action space $\mathcal{A}$ is a discrete set of measurement  positions. The reward function $r$ is defined as:
\begin{equation} \label{eq:5}
    r(s, a, s') = \frac{\Tr(\bm{P^-}_{\mathcal{X_{\mathrm{I}}}}) - \Tr(\bm{P^+}_{\mathcal{X_{\mathrm{I}}}})}{\mathrm{c(\textit{a}^{-}, \textit{a})}}.
\end{equation}
We set $\gamma = 1$, such that $V(\cdot)$ restores \cref{eq:1}.

\subsection{Algorithm Overview} \label{SS:algorithm_overview}

\begin{figure}[!h]
    \centering
    \includegraphics[width=0.48\textwidth, height=2cm]{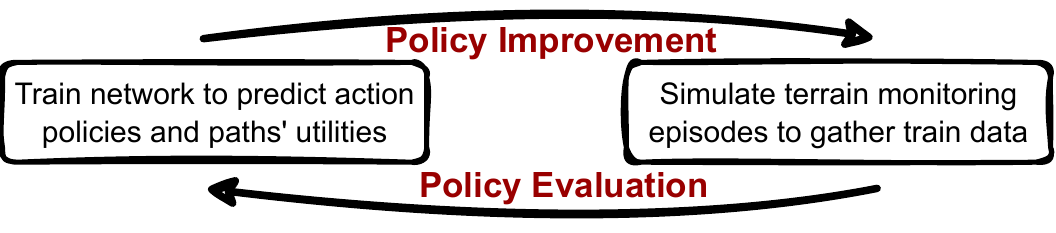}
    \caption{\textbf{Overview of our RL-based IPP approach.} We use a CNN to predict policies and values guiding a tree search to find informative paths for data gathering. The CNN is iteratively trained in simulation to improve the policy and value estimates.}
    \label{F:overview}
\end{figure}

Our goal is to learn the best policies for IPP offline to allow for fast online replanning at deployment. To achieve this, we bring recent advances in RL by \citet{silver2017mastering, silver2018general} into the robotics domain. In a similar spirit, our RL algorithm combines MCTS with a policy-value CNN  (\cref{F:overview}). At train time, the algorithm alternates between episode generation and CNN training. For terrain monitoring, episodes are generated by simulating diverse scenarios varying in map priors, target variables, and initial UAV positions as explained in \cref{SS:episode_generation}. Each episode step from a state $s$ is planned by a tree search producing a target value $V(s)$ given by the simulator and a target policy $\bm{\pi}(s)$ proportional to the tree's root node's action visit counts. $V(s)$ and $\bm{\pi}(s)$ are stored in a replay buffer used for CNN training. As described in \cref{SS:neural_network_architecure}, we introduce a CNN architecture suitable for inference on mobile robots. Further, \cref{SS:low_data_regime} proposes  components for low data regimes in robotics tasks, which are often expensive to simulate.

\subsection{Episode Generation at Train Time} \label{SS:episode_generation}

The most recently trained CNN is used to simulate a fixed number of episodes. An episode terminates when the budget $\mathrm{B}$ is spent or a maximum number of steps is reached. In each step from state $s$, a tree search is executed for a certain number of simulations as described in \cref{SS:mcts_neural_networks}. The policy $\bm{\pi}(s)$ is derived from the root node's action visits $N(s,a)$:
\begin{equation} \label{eq:6}
    \bm{\pi}(s)_{a} = \frac{N(s,a)^{1/\tau}}{\sum_{a' \in \mathcal{A}} N(s, a')^{1/\tau}},
\end{equation}
where $\mathcal{A}$ is the set of next measurement positions reachable within the remaining budget $b$. $\tau$ is a hyper-parameter smoothing policies to be uniform as $\tau \! \to \! \infty$ and collapsing to $\argmax_{a \in \mathcal{A}} \bm{\pi}(s)_{a}$ as $\tau \! \to \! 0$. The action is sampled from $a \sim \bm{\pi}(s)$ and the next map state is given by $\mathcal{KF}(\cdot, \bm{P^{-}}, \cdot, a) = s'_{m}$. $r$ and $b$ are given by \cref{eq:5} and $c(a^{-}, a)$ respectively. The tuple $(s, a, \bm{\pi}(s), V(s), b)$ is stored in the replay buffer.

\subsection{Tree Search with Neural Networks} \label{SS:mcts_neural_networks}

\begin{figure}[!h]
    \centering
    \includegraphics[width=0.48\textwidth]{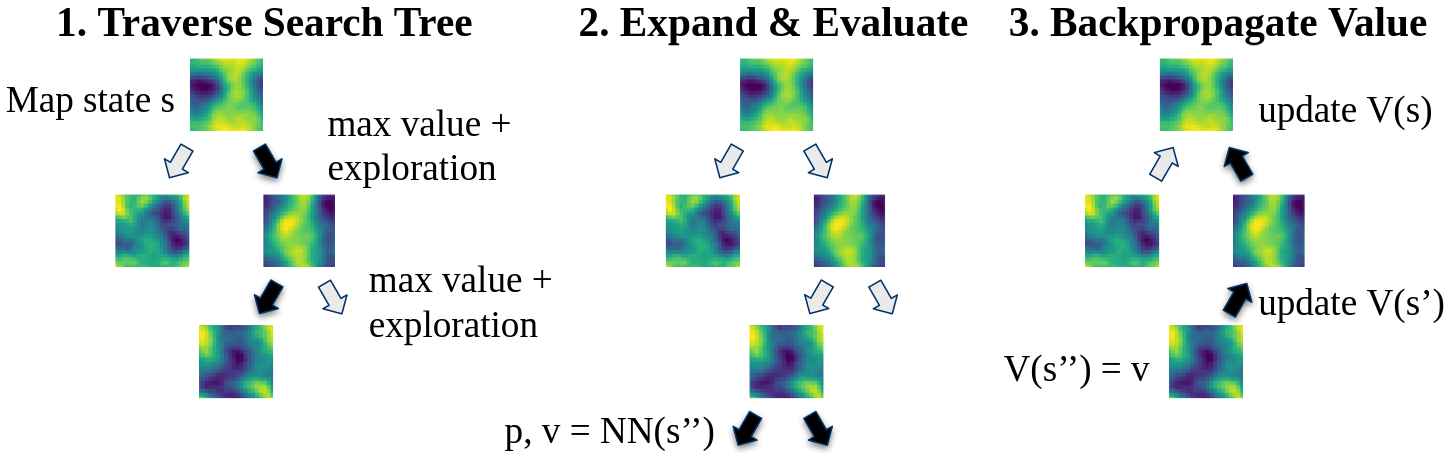}
    \caption{\textbf{Tree search with a CNN}. The colored gradients represent the map state at each node and the arrows indicate how these map states evolve as potential measurement are taken. (1) The predicted policies $\bm{p}$ steer traversing the tree, and (2) the predicted values $v$ avoid expensive high-variance rollouts at leaf nodes. (3) The predicted value is used to update the parent nodes' value estimates.}
    \label{F:mcts_with_neural_nets}
\end{figure}

As shown in \cref{F:mcts_with_neural_nets}, a fixed number of simulations is executed by traversing the tree. Each simulation terminates when the budget or a maximal depth is exceeded. The tree search queries the CNN for policy and value estimates $\bm{p}, v$ at leaf nodes with state $s^{(l)}$ and stores the node's prior probabilities $P(s^{(l)}) = \bm{p}$. The probabilistic upper confidence tree (PUCT) bound is used to traverse the tree \cite{schrittwieser2020mastering}:
\begin{multline} \label{eq:7}
    \mathrm{PUCT}(s, a) = Q(s, a) \\ + ~ P_{a}(s) \frac{\sqrt{N(s)}}{1 + N(s, a)} \left(c_{1} + \log\left[\frac{N(s) + c_{2} + 1}{c_{2}}\right]\right),
\end{multline}
where $Q(s, a) = r(s, a, s') + \gamma \cdot V(s')$ is the state-action value and $N(s) = \sum_{a' \in \mathcal{A}} N(s, a')$ is the visit count of the parent node $s$. $c_{1}, c_{2} \in \mathbb{R}^{+}$ are exploration factors. We choose the next action $a = \argmax_{a' \in \mathcal{A}} \mathrm{PUCT(s, \textit{a'})}$.

As the reward has no fixed scale, in \cref{eq:7}, we min-max normalize $Q(s,a)$ to $\Tilde{Q}(s,a) \in [0, 1]$ across all $\textit{a} \in \mathcal{A}$ in $s$. To enforce exploration, similarly to \citet{silver2017mastering}, we add Dirichlet-noise to $P(s^{(r)})$ of the root node with state $s^{(r)}$:
\begin{equation} \label{eq:9}
    P(s^{(r)}) = (1 - \epsilon) \bm{\pi}(s^{(r)}) + \epsilon \bm{\eta},
\end{equation}
where $\epsilon \in [0, 1]$, and the noise $\bm{\eta} \sim \mathrm{Dir}(\delta)$, $\delta > 0$.

\subsection{Network Architecture \& Training} \label{SS:neural_network_architecure}

The CNN $f_{\theta}(s) = (\bm{p}, v)$ is parameterized by $\theta$ predicting a policy $\bm{p}$ and value $v$. Input to the CNN are (a) the min-max normalized current map covariance $\bm{P^-}_{\mathcal{X_{\mathrm{I}}}}$; (b) the remaining budget $b \leq \mathrm{B}$ normalized over $\mathrm{B}$; (c) the UAV position $a^{-}$ normalized over the bounds of the 3D action space $\mathcal{A}$; and (d) a costs feature map $\mathrm{C}$ of same shape as $\bm{P^-}_{\mathcal{X_{\mathrm{I}}}}$ with $\mathrm{C}[i, :] = c(a_{i}, a)$, subsequently min-max normalized. Note that the scalar inputs are expanded to feature maps of the same shape as $\bm{P^-}_{\mathcal{X_{\mathrm{I}}}}$. Additionally, we input a history of the previous two covariance, position, and budget input planes.

\begin{figure}[!h]
    \centering
    \includegraphics[width=0.48\textwidth]{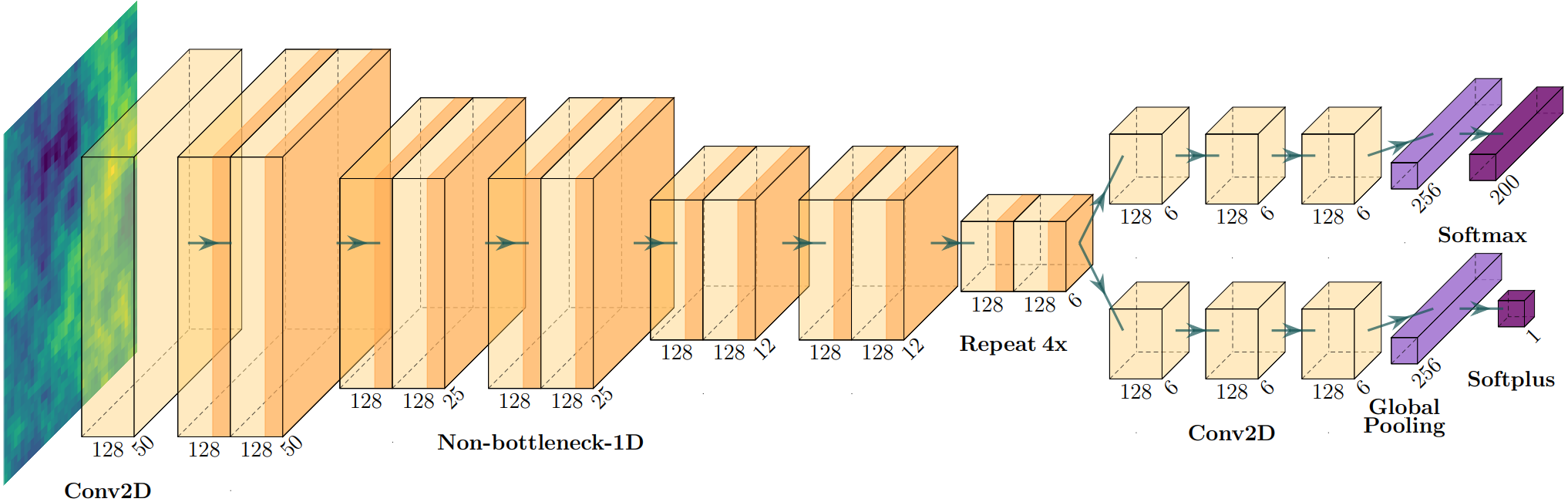}
    \caption{\textbf{Our policy-value CNN architecture.} We leverage an ERFNet encoder \cite{romera2017erfnet} with $10$ residual blocks providing shared representations for the policy and value prediction. Both heads consist of three convolutional blocks and global average pooling to make the CNN input size-agnostic. Last, fully connected layers project to a policy vector (Softmax) and single positive scalar value (Softplus). Feature map dimensions are w.r.t. $10 \! \times \! 10$ grid maps $\mathcal{X}$.}
    \label{F:policy_value_network}
\end{figure}

As visualized in \cref{F:policy_value_network}, the CNN has a shared encoder. We leverage Non-bottleneck-1D blocks proposed by \citet{romera2017erfnet} to reduce inference time. The encoder is followed by two separate prediction heads for policy and value estimates. Both heads consist of three blocks with 2D convolution, batch norm, and SiLU activations. The last block's output feature maps in each head are flattened to fixed dimensions by global average and max pooling before applying a fully connected layer. This reduces the number of parameters and ensures an input size-agnostic architecture.
The CNN parameters $\theta$ are trained with stochastic gradient descent (SGD) on mini-batches of size $96$ to minimize:
\begin{equation} \label{eq:10}
    l(s) = \alpha \cdot (V(s) - v)^{2}  - \beta \cdot \bm{\pi}(s)^{T} \log \bm{p} +  \lambda \cdot {\lVert \theta \rVert}^{2},
\end{equation}
where the loss coefficients $\alpha, \beta, \lambda \geq 0$ are hyperparameters. SGD uses a one-cycle learning rate over three epochs \cite{smith2018disciplined}.

\begin{table*}[h]
\centering
\setlength{\tabcolsep}{2.7pt}
\begin{tabular}{@{}lccccccc@{}}
Variant                           & 33\% $\Tr(\bm{P})$ $\downarrow$ & 67\% $\Tr(\bm{P})$ $\downarrow$ & 100\% $\Tr(\bm{P})$ $\downarrow$ & 33\% RMSE $\downarrow$ & 67\% RMSE $\downarrow$ & 100\% RMSE $\downarrow$ & Runtime {[}s{]} $\downarrow$ \\ \midrule
Baseline as in \cref{S:approach}        & \textbf{73.61}                   & 31.83                   & \textbf{12.44}                   & \textbf{0.15}                  & 0.09                  & \textbf{0.05}                  & 0.64                         \\
(i) w/ fixed off-policy window             & 83.25                   & 50.17                   & 24.68                   & 0.16                  & 0.12                 & 0.09                  & 0.68                         \\
(i) w/ fixed exploration constants       & 95.27                   & 39.46                   & 21.86                   & 0.20                  & 0.11                  & 0.08                  & 0.65                         \\
(i) w/o forced playouts + policy pruning & 79.23                   & \textbf{28.53}                   & 22.62                   & 0.18                  & 0.09                  & 0.07                  & 0.66                         \\
(ii) w/o global pooling bias blocks       & 103.58                   & 45.78                   & 31.44                   & 0.19                  & 0.11                  & 0.10                  & 0.64                         \\
(ii) 5 residual blocks in encoder          & 82.90                   & 29.94                   & 17.94                   & 0.16                  & \textbf{0.08}                  & 0.07                  & \textbf{0.55}                         \\
(ii) w/o input feature history            & 102.40                   & 40.48                   & 31.33                   & 0.20                  & 0.10                  & 0.09                  & 0.66                         \\ 
\end{tabular}
\caption{\textbf{Ablation study results}. We systematically (i) remove components, and (ii) change the CNN architecture to quantify their impact. Remaining uncertainty $\Tr(\bm{P})$ and RMSE in the map state are evaluated after 33\%, 67\%, and 100\% spent effective mission time. 
Our approach as proposed in \cref{S:approach} achieves the fastest and most stable reductions in uncertainty and RMSE over the mission time.
} \label{T:results_ablation}
\end{table*}

\subsection{AlphaZero in Low Data Regimes} \label{SS:low_data_regime}

\textit{Adaptive} IPP with spatio-temporal correlations is expensive to simulate. Opposed to fast-to-simulate games such as Go or chess \cite{silver2018general}, real-world robotics tasks are often limited in the number of simulations and episodes at train time.

A major shortcoming of the original AlphaZero algorithm \cite{silver2018general} is that the policy targets in \cref{eq:6} merely reflect the tree search exploration dynamics. However, the raw action visit counts $N(s,a)$ do not necessarily capture the gathered state-action value information for a finite number of simulations. Hence, with only a moderate number of simulations per episode step, AlphaZero tends to overemphasize initially explored actions in subsequent training iterations, 
leading to bias in training data and thus overestimated state-action values. As \cref{eq:7} is also guided by $Q(s,a)$, the overemphasis on initially explored actions induces overfitting and low-quality policies. Next, we introduce methods to solve these problems and increase efficiency of our RL algorithm.

To avoid overemphasizing random actions in the node selection, a large exploration constant $c_{1}$ in \cref{eq:7} is desirable. However, in later training iterations, increasing exploitation of known good actions is required to ensure convergence. Thus, we propose an exponentially decaying exploration constant $c_{1}^{(i)} = \max(c_{1}^{(start)} \cdot \lambda_{c_{1}}^{i}, ~ c_{1}^{(min)})$, where $i \in \mathbb{N}$ is the training iteration number, $c_{1}^{(start)} \! > \! 0$ is the initial constant, $\lambda_{c_{1}} \! > \! 0$ is the exponential decay factor, and $c_{1}^{(min)} \! > \! 0$ is the minimal value.
Similarly, for the Dirichlet exploration noise in \cref{eq:9} defined by $\delta$, we introduce an exponentially decaying scheduling $\delta^{(i)} = \max(\delta^{(start)} \cdot \lambda_{\delta}^{i}, ~ \delta^{(min)})$, where $\delta^{(start)} \! > \! 0$ is the initial value, $\lambda_{\delta} \! > \! 0$ is the exponential decay factor, and $\delta^{(min)} \! > \! 0$ is the minimal value. A high $\delta^{(start)}$ around $1$ leads to a uniform noise distribution $\bm{\eta}$ avoiding overemphasis on random actions. However, $\delta^{(i)}$ should decrease with increasing $i$ to exploit the learned $\bm{\pi}(s)$.

Next, we propose an increasing replay buffer size $w$ to accelerate training. Similar to our approach, adaptive replay buffers are known to improve performance in other RL domains \cite{liu2018effects}. On the one hand, a substantial amount of data is required to train the CNN on a variety of paths. On the other hand, the loss (\cref{eq:10}) initially shows sudden drops when outdated train data leaves the replay buffer. Thus, in early training stages, a small $w$ improves convergence speed. Larger $w$ in later training stages help regularize training and ensure train data diversity. Hence, $w$ is adaptively set to:
\begin{equation} \label{eq:13}
    w^{(i)} = \min\left(\left\lfloor w^{(start)} + \frac{i}{w^{(step)}} \right\rfloor, w^{(max)}\right),
\end{equation}
where $w^{(start)} \! \in \! \mathbb{N}^{+}$ and $w^{(max)} \! \in \! \mathbb{N}^{+}$ are the initial and maximum window sizes. The window size is increased by one each $w^{(step)} \! \in \! \mathbb{N}^{+}$ training iterations.

Moreover, we adapt two techniques introduced by \citet{wu2019accelerating} for the game of Go. First, forced playouts and policy pruning decouple exploration dynamics and policy targets. While traversing the search tree, underexplored root node actions $a$ are chosen by setting $\mathrm{PUCT}(s^{(root)}, a) \! = \! \infty$ in \cref{eq:7}. In \cref{eq:6}, action visits $N(s^{(root)}, a)$ are subtracted unless action $a$ led to a high-value successor state.
%
%
Second, in regular intervals in the encoder, and as the first layers of the value and policy head, we use global pooling bias layers. 
This enables our CNN to focus on local and global features required for IPP. Further details are discussed by \citet{wu2019accelerating}. 

\subsection{Planning at Mission Time} \label{SS:planning_at_deploy_time}

Replanning is performed after each map update. Using the trained CNN, we perform tree search from the current state $s$, choosing action $a$ from $\bm{\pi}(s)$ with $\tau \! \to \! 0$ in \cref{eq:6}, i.e. $a = \argmax_{a \in \mathcal{A}} \bm{\pi(s)}_{a}$. Since $\bm{\pi}(\cdot)$ steers the exploration, the tree search is highly sample-efficient. This way, the number of tree search simulations is greatly reduced to allow fast replanning. Note that policy noise injection and forced playouts at the root are disabled to improve performance.

\begin{figure*}[!h]
    \centering
    \includegraphics[width=0.245\textwidth]{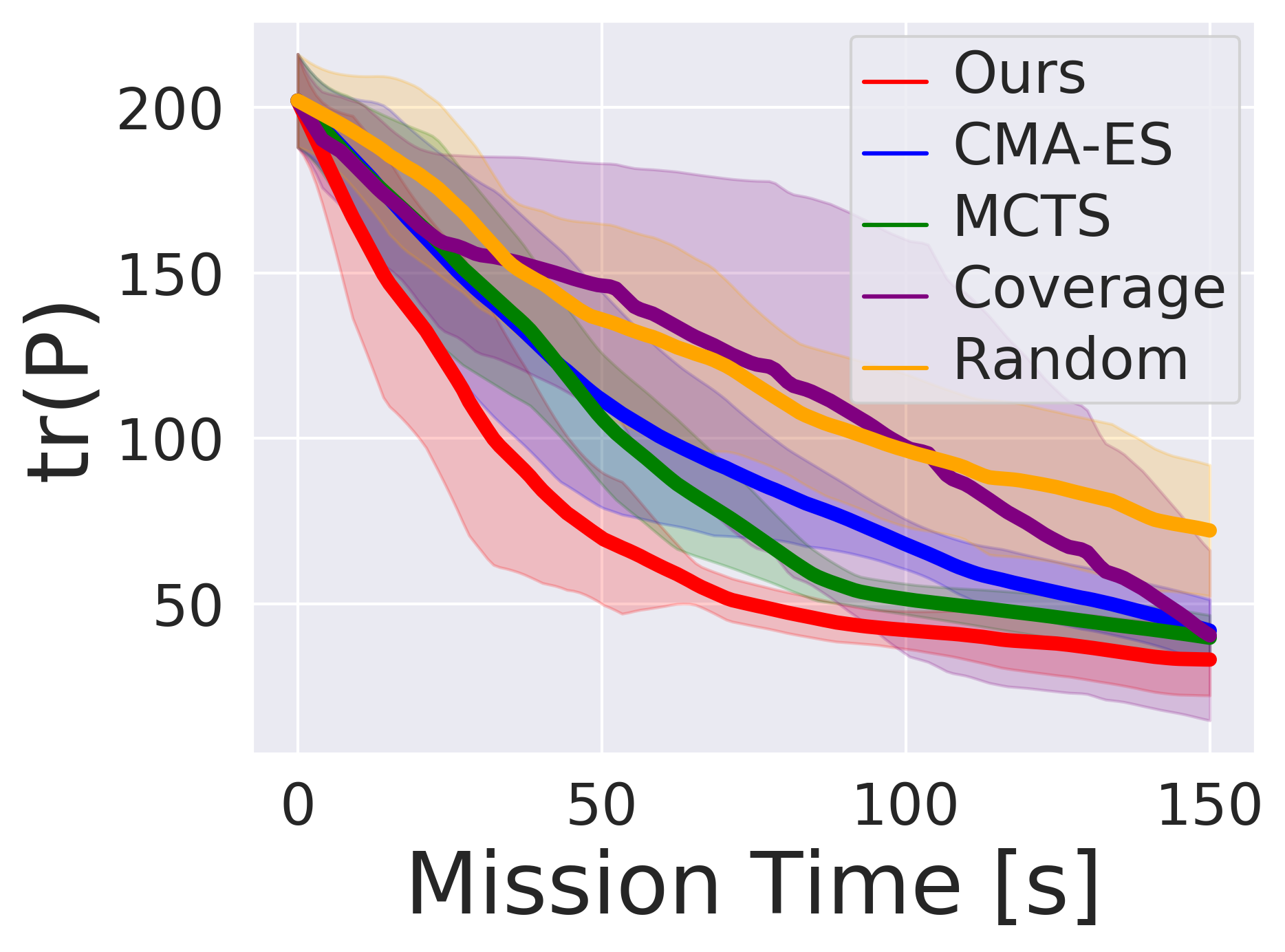}
    \includegraphics[width=0.245\textwidth]{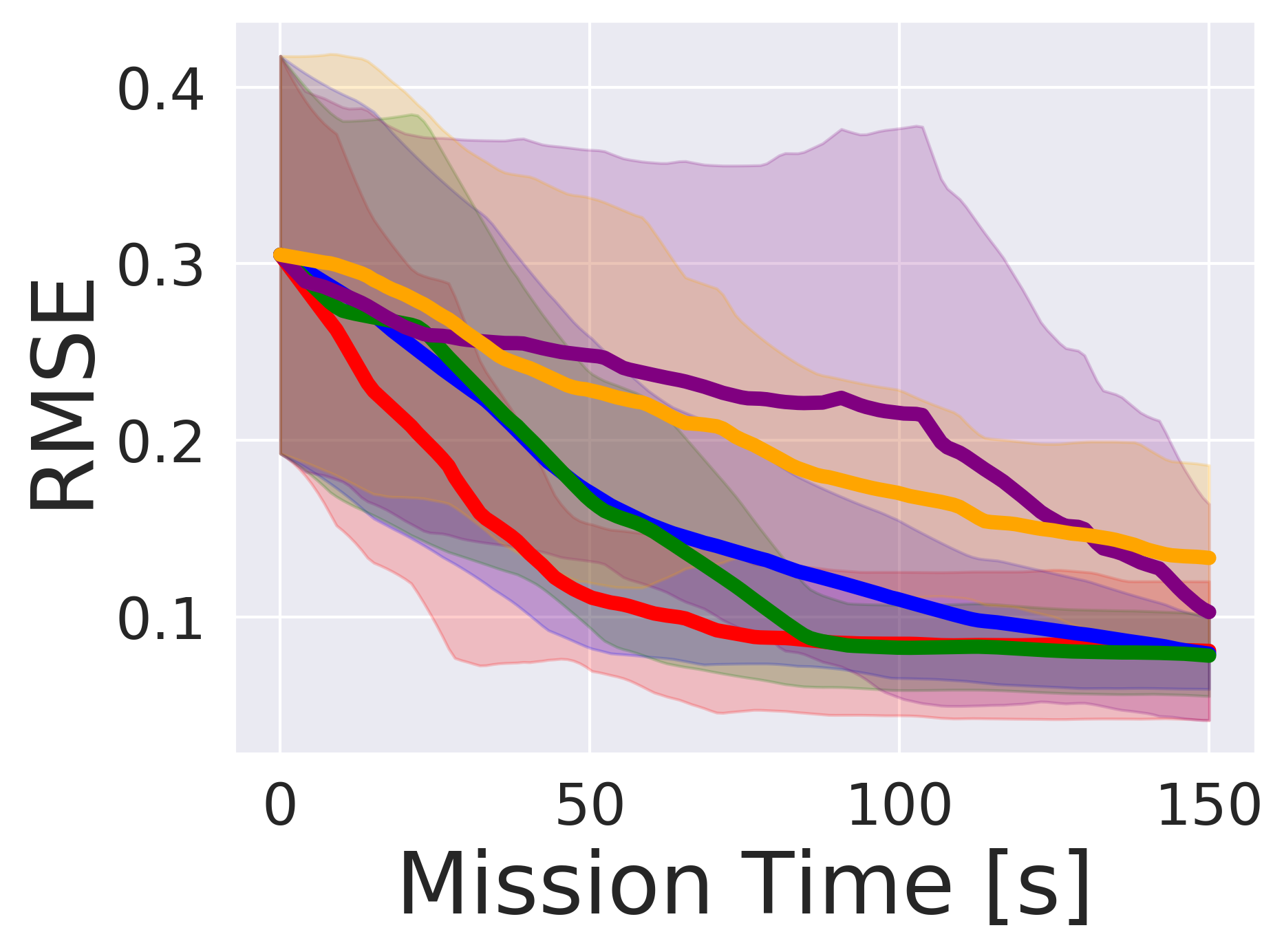}
    \includegraphics[width=0.245\textwidth]{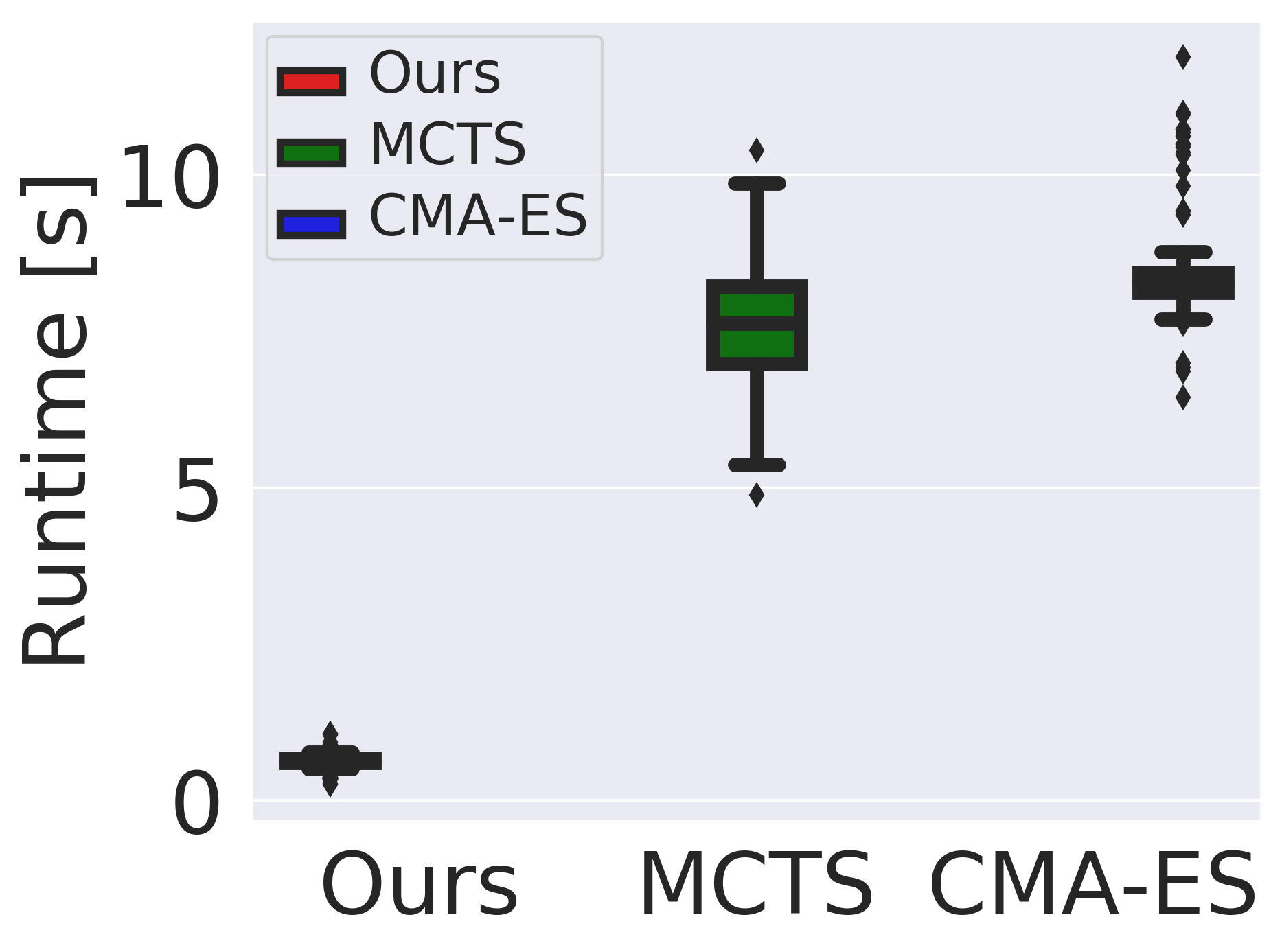}
    \includegraphics[width=0.22\textwidth]{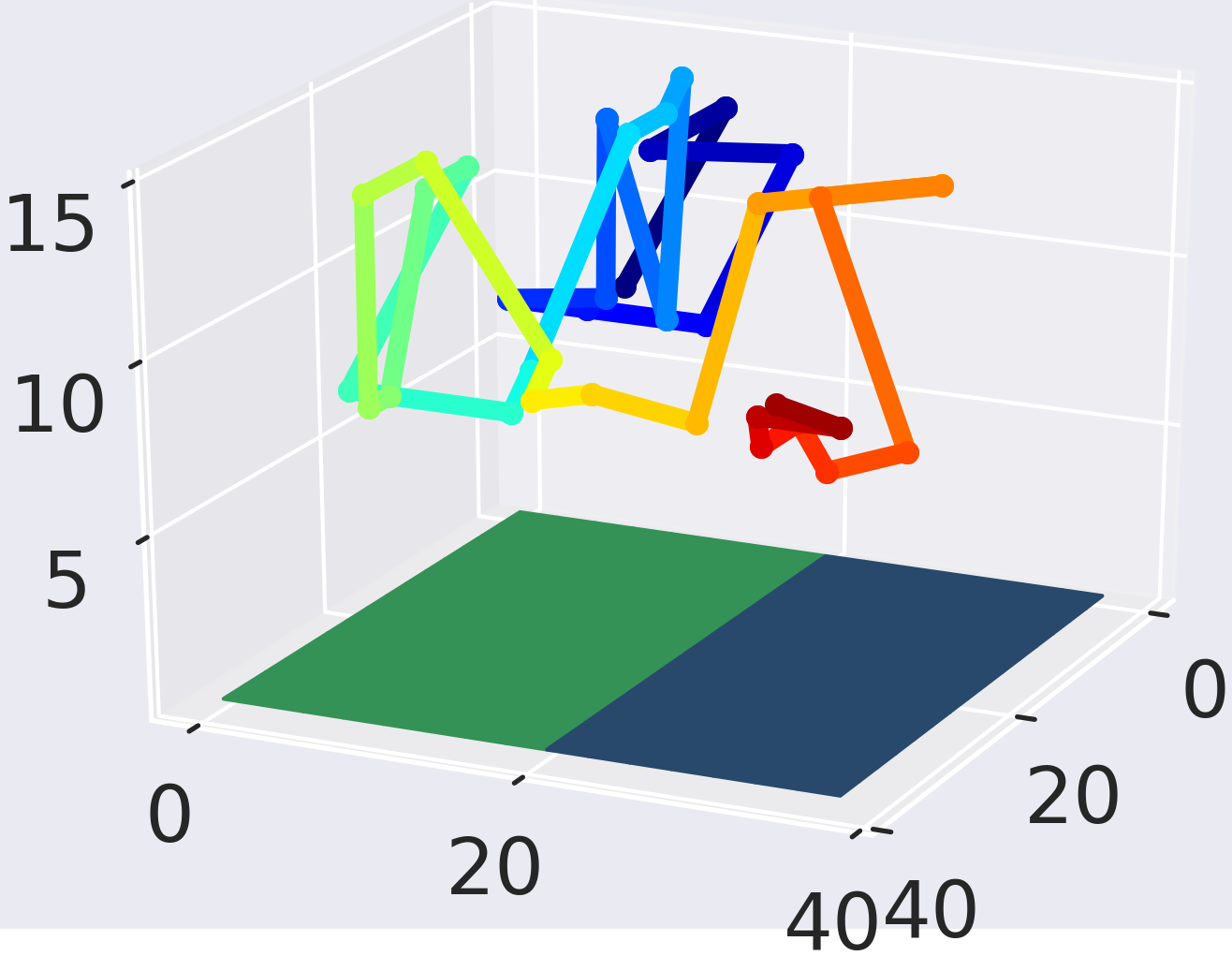}
\caption{\textbf{Evaluation of our approach against benchmarks.} On average, our RL approach ensures the fastest uncertainty and RMSE reduction in regions of interest over mission time. Solid lines indicate means over 10 trials, and shaded regions indicate the standard deviations. Note that there is inherent variability due to the randomly generated hotspot locations. However, our method ensures stable performance over changing environments. Further, replanning runtime is reduced by a factor of $8-10 \times$. The planned path (evolving over time from blue to red) validates the \textit{adaptive} behavior of our approach exploring the terrain with focus on the high-value region (green).} \label{F:benchmark_comparison}
\end{figure*}

\section{Experimental Results} \label{S:results}

This section presents our experimental results. We first validate our RL approach in an ablation study, then assess its IPP and runtime performance in terrain monitoring scenarios.

\subsection{Experimental Setup} \label{SS:experiment_setup}

Our simulation setup considers terrains $\xi$ with 2D discrete field maps $\mathcal{X}$ with values between 0 and 1, randomly split in high- and low-value regions to create regions of interest as defined by \cref{eq:4}. We model ground truth terrain maps of $40 \times 40 ~ m$ and $r \times r ~ [m], r \in \mathbb{R},$ resolution. The UAV action space of measurement locations is defined by a discrete 3D lattice above the terrain $\xi$. The lattice mirrors the $g \times g$ grid map $\mathcal{X}$ on two altitude levels ($8 ~ m$ and $14 ~ m$), resulting in $2 \cdot g^{2}$ actions. The missions are implemented in Python on a desktop with a 1.8 GHz Intel i7, 16 GB RAM without GPU acceleration to avoid unfair advantages in inference speed of our CNN. We repeat the missions $10$ times and report means and standard deviations. Our RL algorithm is trained offline on a single machine with a 2.2 GHz AMD Ryzen 9 3900X, 63GB RAM, and a NVIDIA GeForce RTX 2080 Ti GPU.

We use the same altitude-dependent inverse sensor model as \citet{popovic2020informative} to simulate camera measurement noise, assuming a downwards-facing square camera footprint with $60 \degree$ FoV. The prior map mean is uniform with a value of $0.5$. The GP is defined by Mat\'ern $3/2$ kernel with length scale $3.67$, signal variance $1.82$, and noise variance $1.42$ by maximizing log marginal likelihood over independent maps. The threshold $\mu_{th} = 0.4$ defines regions of interest.

We set the mission budget $B = 150 ~ s$, the UAV initial position to $(2, 2, 14) ~ m$, the acceleration-deceleration $\pm u_{a} = 2 ~ m/s^2$ with maximum speed $u_{v} = 2 ~ m/s$. At train time, each episode randomly generates a new ground truth map, map priors from a wide range of GP hyperparameters and UAV start positions, such that our approach has no unfair overfitting advantage. We evaluate map uncertainty with $\Tr(\bm{P}^{+})$ and the root mean squared error (RMSE) of $\bm{\mu}^{+}$ in regions of interest to assess planning performance. Lower values indicate better performance. In contrast to earlier work \cite{popovic2020informative, choudhury2020adaptive, hitz2017adaptive}, the remaining budget does not only incorporate the path travel time, but also the planning runtime, as relevant for robotic platforms with limited on-board resources. We refer to the spent budget $B$ as the \textit{effective mission time}.

\subsection{Ablation Study} \label{SS:ablation_study}

This section validates the algorithm design and CNN architecture introduced in \cref{S:approach}. We perform an ablation study comparing our approach to versions of itself (i) removing proposed training procedure components, and (ii) changing the CNN architecture. We assume a resolution of $r\!=\!4 ~ m$ resulting in a $10\!\times\!10$ grid map $\mathcal{X}$ with $\mathcal{A}$ of $200$ actions. Note that the results do not depend on the actual size of $\mathcal{X}$ and $\mathcal{A}$. We generate a small number of $280$ episodes in each iteration and terminate training after $40$ iterations. Each tree search is executed as described in \cref{SS:planning_at_deploy_time} with $10$ simulations and exploration constants $c_{1}^{(start)}\!=\!15, c_{1}^{(min)}\!=\!4, \lambda_{c_{1}}\!=\!\lambda_{\delta}\!=\!0.8, \delta^{(start)}\!=\!1.0, \delta^{(min)}\!=\!0.3, c_{2}\!=\!10000$. \cref{T:results_ablation} summarizes our results. We evaluate map uncertainty and RMSE over the posterior map state after 33\%, 67\%, and 100\% \textit{effective mission time}, and average planning runtime.

As proposed in \cref{eq:13}, the training procedure considers a replay buffer with adaptive size $w^{(start)}\!=\!1, w^{(step)}\!=\!2, w^{(max)}\!=\!10$. Convergence speed, and thus performance, improves compared to a fixed-size buffer $w^{(i)}\!=\!10$. 
Also, our proposed exploration constants scheduling scheme improves plan quality by stabilizing the exploration-exploitation trade-off compared to fixed constants $c_{1}^{(i)}\!=\!4, \delta^{(i)}\!=\!0.3$. Further, the results show the benefits of including a history of the two previous map states and UAV positions in addition to their current values. Interestingly, reducing the encoder depth from $10$ to $5$ blocks and removing forced playouts both perform reasonably well, but still lead to worse results in later mission stages. This suggests that deeper CNNs and forced playouts facilitate learning in larger grid maps and longer missions. Similarly, global pooling bias blocks help learning global map features, which benefits information-gathering.

\subsection{Comparison Against Benchmarks} \label{SS:simulation_results}

Next, our RL algorithm is evaluated against benchmarks. We set a resolution $r\!=\!2.5 ~ m$, hence $\mathcal{X}$ is a $15\!\times\!15$ grid, and $\mathcal{A}$ has $450$ actions. Our approach is compared against: (a) uniform random sampling in $\mathcal{A}$; (b) coverage path with equispaced measurements at a fixed $8 ~ m$ altitude; (c) MCTS with progressive widening \cite{sunberg2018online} for large action spaces and a generalized cost-benefit rollout policy proposed by \citet{choudhury2020adaptive}; (d) a state-of-the-art IPP framework using the Covariance Matrix Adaptation Evolution Strategy (CMA-ES) proposed by \citet{popovic2020informative}. All planners consider a $5$-step horizon. We set CMA-ES parameters to $45$ iterations, $12$ offsprings, and $(4,4,3)~m$ coordinate-wise step size to trade-off between performance and runtime. The permissible next actions of MCTS are reduced to a radius of $11 ~ m$ around the UAV to be computable online, resulting in $\sim 115$ next actions per move. For a fair comparison, we trained our approach on this restricted action space, which is still much larger than studied in prior work \citep{choudhury2020adaptive,popovic2020localisation}.

\cref{F:benchmark_comparison} reports the results obtained using each approach.
Our method substantially reduces runtime, achieving a speedup of $8-10 \times$ compared to CMA-ES and MCTS. This result highlights the improved sample-efficiency in our tree search and confirms that the CNN can learn informative actions from training in diverse simulated missions.
Random sampling performs poorly as it reduces uncertainty and RMSE in high-value regions only by chance. The coverage path shows high variability since data-gathering efficiency greatly depends on the problem setup, i.e. hotspot locations relative to the preplanned path. Our approach outperforms this benchmark with much greater consistency.

\subsection{Temperature Mapping Scenario} \label{SS:real_world_data_experiment}

We demonstrate our RL-based IPP approach in a photorealistic simulation using real-world surface temperature data. The data was collected in a $40\!\times\!40 ~ m$ crop field nearby Forschungszentrum J\"{u}lich, Germany $(50.87\degree\,\textrm{lat.}, 6.44\degree\, \textrm{lon.})$ on July 20, 2021 with a DJI Matrice 600 UAV carrying a Vue Pro R 640 thermal sensor. The UAV executed a coverage path at $100 ~ m$ altitude to collect images, which were then processed using Pix4D software to generate an orthomosaic representing the target terrain in our simulation as depicted in \cref{F:teaser}-left. The aim is to validate our method for adaptively mapping high-temperature areas in this realistic setting.

\begin{figure}[!h]
    \centering
    \includegraphics[width=0.1999\textwidth]{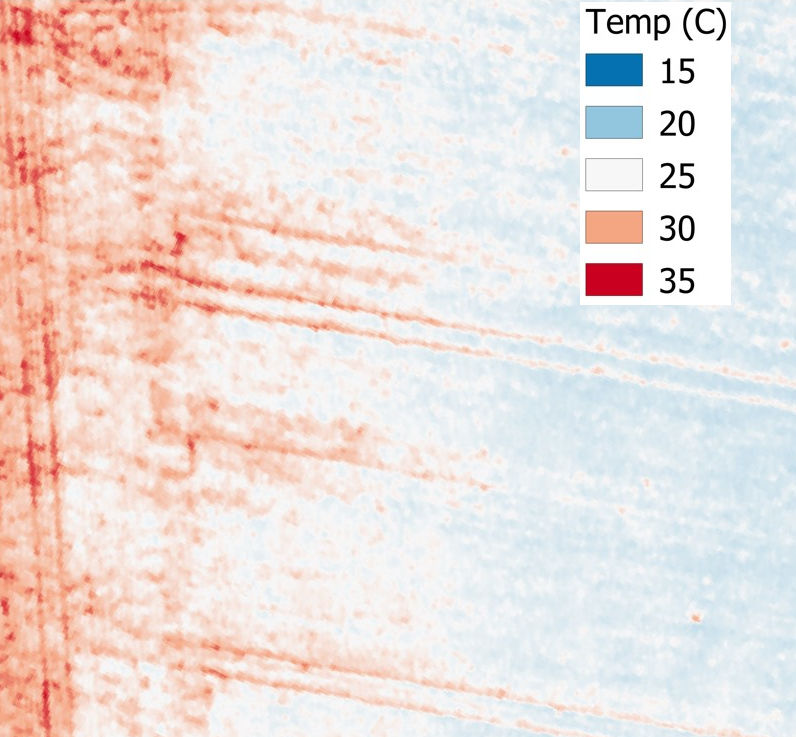}
    \includegraphics[width=0.2599\textwidth]{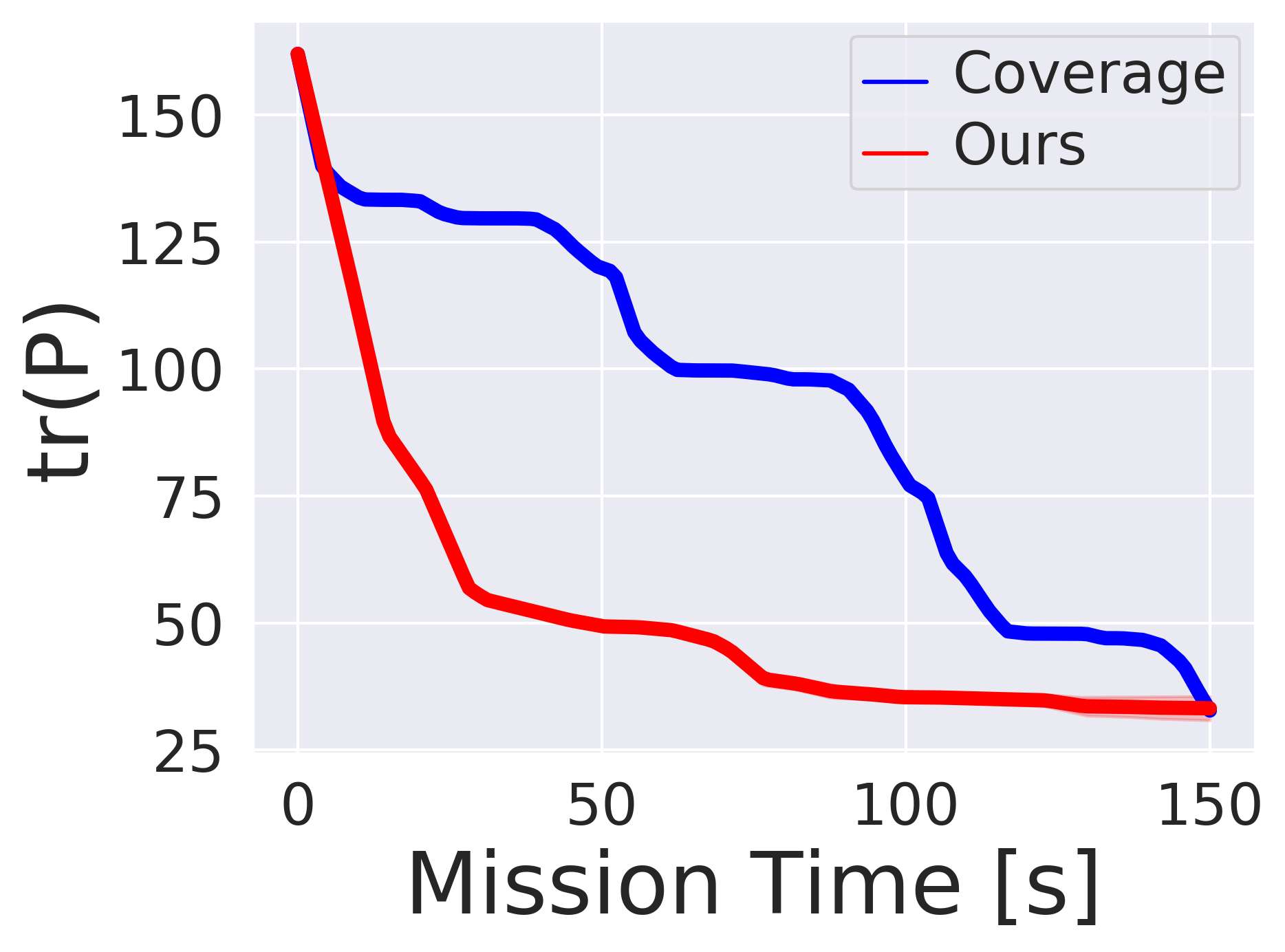}
    \caption{\textbf{Real-world scenario.} (Left) Surface temperature of a crop field used for the conducted real-world experiments. (Right) Our RL approach ensures fast uncertainty reduction in high-temperature regions (red) outperforming traditionally used coverage paths.}
    \label{F:real_world_results}
\end{figure}
For fusing new data into the map, we assume altitude-dependent sensor noise as described in \cref{SS:experiment_setup}. The terrain is discretized using a uniform $2.5 ~ m$ resolution.
We compare our RL-based online algorithm against a fixed $8 ~ m$ altitude lawnmower path as a traditional baseline. Our approach is trained only on synthetic simulated data as shown in \cref{F:benchmark_comparison}.

\cref{F:teaser}-right shows the planned 3D path above the terrain using our strategy. This confirms that our method collects targeted measurements in hotter areas of interest (red) by efficient online replanning. This is reflected quantitatively in \cref{F:real_world_results}-right as our approach ensures fast uncertainty reduction while a coverage path performs worse as it cannot adapt mapping behaviour. These results verify the successful transfer of our model trained in simulation to real-world data and demonstrate its benefits over a traditional approach.


\section{Conclusions and Future Work} \label{S:conclusions}

This paper proposes a new RL-based approach for online adaptive IPP using resource-constrained robots. The algorithm enables sample-efficient planning in large action spaces and high-dimensional state spaces, enabling fast information gathering in active sensing tasks. A key feature of our approach are components for accelerated learning in low data regimes. We validate the approach in an ablation study, and evaluate its performance compared to multiple benchmarks. Results show that our approach drastically reduces planning runtime, enabling efficient adaptive replanning.
Future work will investigate extending our algorithm to multi-robot teams and dynamically growing maps. We plan to conduct real-world field experiments to validate our method.


\section*{Acknowledgement}

We would like to thank Jordan Bates  from Forschungszentrum J\"{u}lich for providing the real-world data.

\bibliographystyle{IEEEtranN}
\footnotesize
\bibliography{2022-icra-rueckin}

\end{document}